\documentclass[11pt,a4paper]{article}
\usepackage[hyperref]{naaclhlt2019}
\usepackage{times}
\usepackage{latexsym}
\usepackage{lipsum}
\usepackage{booktabs}
\usepackage{amsmath}
\usepackage{bm}
\usepackage{graphicx}
\usepackage{subcaption}
\usepackage{todonotes}
\usepackage[capitalize,nameinlink]{cleveref}
\usepackage{color,soul}
\usepackage[normalem]{ulem}
\usepackage{enumitem}

\usepackage{url}

\aclfinalcopy % Uncomment this line for the final submission
\title{Adversarial Regularization for Visual Question Answering:\\Strengths, Shortcomings, and Side Effects}

\author{\normalfont{\textbf{Gabriel Grand}\textsuperscript{1} and \textbf{Yonatan Belinkov}\textsuperscript{1,2}} \\
\textsuperscript{1}Harvard John A. Paulson School of Engineering and Applied Sciences\\
\textsuperscript{2}MIT Computer Science and Artificial Intelligence Laboratory\\
Cambridge, MA, USA\\
{\tt ggrand@alumni.harvard.edu, belinkov@seas.harvard.edu}}

\date{}

\begin{document}
\maketitle
\begin{abstract}
%\parskip 0pt
% \vspace{-2pt}

Visual question answering (VQA) models have been shown to over-rely on linguistic biases in VQA datasets, answering questions ``blindly'' without considering visual context. Adversarial regularization (\mbox{AdvReg}) aims to address this issue via an adversary sub-network that encourages the main model to learn a bias-free representation of the question.
In this work, we investigate the strengths and shortcomings of \mbox{AdvReg} with the goal of better understanding how it affects inference in VQA models. Despite achieving a new state-of-the-art on VQA-CP, we find that \mbox{AdvReg} yields several undesirable side-effects, including unstable gradients and sharply reduced performance on in-domain examples. We demonstrate that gradual introduction of regularization during training helps to alleviate, but not completely solve, these issues.
Through error analyses, we observe that \mbox{AdvReg} improves generalization to binary questions, but impairs performance on questions with heterogeneous answer distributions. Qualitatively, we also find that regularized models tend to over-rely on visual features, while ignoring important linguistic cues in the question. Our results suggest that \mbox{AdvReg} requires further refinement before it can be considered a viable bias mitigation technique for VQA.

\end{abstract}

\section{Introduction}
%\vspace{-2pt}

% In recent years, the Visual Question Answering (VQA) community has uncovered a cluster of bias-related issues that act as potential confounds to VQA research.
% In recent years, the Visual Question Answering (VQA) community has identified bias as a significant confound that has a broad, adverse effects on research progress.
% In recent years, the Visual Question Answering (VQA) community has grown increasingly conscious of the broad, adverse effects of bias on research progress.
% In recent years, the Visual Question Answering (VQA) community has uncovered a cluster of bias-related issues that undermine VQA research progress.
In recent years, the Visual Question Answering (VQA) community has grown increasingly cognizant of the confounding role that bias plays in VQA research. Many popular VQA datasets have been shown to contain systematic language biases that enable models to cheat by answering questions ``blindly'' without considering visual context \citep{agrawal2016analyzing, zhang2016yin, VQA2, VQA-CP}.

\pagebreak

Efforts to address this problem have mainly focused on constructing more balanced datasets \citep{zhang2016yin, VQA2, johnson2017clevr, chao2018being}. However, any benchmark that involves crowdsourced data is likely to encode certain cognitive and/or social biases \citep{van2016stereotyping, misra2016seeing, eickhoff2018cognitive}.
An alternate approach is to develop models that can generalize to novel domains with different biases. In this spirit, \citet{VQA-CP} introduced VQA under Changing Priors (VQA-CP), a new benchmark in which the distribution of answers varies significantly between train and test splits. Existing models, which tend to rely heavily on the distribution of answers in the training set, perform poorly on VQA-CP \citep{VQA-CP}.

One approach to mitigating bias that has recently gained interest is a technique called adversarial regularization (\mbox{AdvReg}). In \mbox{AdvReg}, an adversary sub-network performs an inference task based on a subset of the input features; in this case, the adversary attempts to predict answers based only on the question. Successful performance by the adversary indicates that the main network has learned a biased input representation. Negated gradient updates from the adversary are backpropagated to a shared encoder to encourage the main network to learn a bias-neutral representation of the question. Recently, \citet{ramakrishnan2018overcoming} applied \mbox{AdvReg} to VQA and found that it improves generalization to out-of-domain examples on VQA-CP test.
 
Despite this initial success, \mbox{AdvReg} is still a relatively new methodology, and its effects on representation learning in neural networks remain largely unknown. In this study, we explore \mbox{AdvReg} with the goal of better understanding how this technique affects inference in VQA models. We apply \mbox{AdvReg} to the Pythia VQA architecture \citep{pythia18arxiv}, achieving a new state-of-the-art on VQA-CP v1 and v2.
However, we find that \mbox{AdvReg} yields a number of previously unreported and undesirable side-effects. We first observe that \mbox{AdvReg} introduces significant noise into gradient updates that creates instability during training. This finding motivates the introduction of a new scheduling technique that gradually introduces regularization over the course of training. We find that scheduling improves gradient stability in the early phases of adversarial training and improves performance on VQA-CP v2. However, even with scheduling, \mbox{AdvReg} significantly reduces performance on in-domain examples. This side-effect suggests that like many statistical regularization methods, \mbox{AdvReg} offers a trade-off between in-domain and out-of-domain performance.

To investigate the strengths and weaknesses of regularized models, we perform quantitative and qualitative error analyses. We find that \mbox{AdvReg} is especially helpful with Yes/No questions, but reduces performance on questions with heterogeneous answers. We also visualize a number of successes and failures of \mbox{AdvReg}, revealing that regularized models often ignore linguistic cues in the question and are heavily swayed by salient visual features. These findings suggest an under-utilization of key information in the question.

The contributions of this work are two-fold. First, we share practical tips for dealing with the idiosyncrasies of \mbox{AdvReg}. Second, we highlight some core drawbacks of \mbox{AdvReg} that have not previously been reported in the literature. By drawing attention to these shortcomings, we hope to motivate future efforts to refine \mbox{AdvReg}.

%\vspace{-2pt}
\section{Related Work}
%\vspace{-2pt}

% \vspace{-2pt}
\paragraph{Biases in VQA datasets}
A growing body of work points to the existence of biases in popular VQA datasets \citep{agrawal2016analyzing, zhang2016yin, jabri2016revisiting, VQA2, johnson2017clevr, chao2018being, VQA-CP, thomason2018shifting}. In VQA v1~\citep{VQA1}, for instance, for questions of the form, ``What sport is...?'', the correct answer is ``tennis'' 41\% of the time, and for questions beginning with ``Do you see a...?'' the correct answer is``yes'' 87\% of the time \citep{zhang2016yin}. By exploiting these biases, models can disregard the image and still achieve high VQA scores.

\pagebreak
\paragraph{Biases in other language tasks} Language biases have also been reported in natural language inference (NLI)~\citep{gururangan-EtAl:2018:N18-2,1804.08117,poliak-EtAl:2018:S18-2}, reading comprehension~\citep{kaushik2018much}, and story cloze completion~\citep{schwartz2017effect}. Many of these tasks are concerned with inferring the relationship between two objects. As in VQA, models can often succeed by learning biases associated with one of these objects, while ignoring the other. 

\vspace{-2pt}
\paragraph{Biases in other vision tasks} Images can also encode certain associative biases. For instance, the Commmon Objects in Context (COCO) image dataset \citep{lin2014coco}, which is used in VQA, has been shown to contain prominent gender biases \citep{zhao2017men, burns2018women}. Recently, \citet{burns2018women} introduced a technique that encourages the assignment of equal gender probability when gender information is occluded from an image. Their Appearance Confusion Loss can be viewed as a vision captioning analogue to AdvReg for VQA.

\vspace{-2pt}
\paragraph{Mitigating bias}
Initial efforts to address bias in VQA focused on debiasing existing datasets. VQA v2 introduced complimentary examples with different answers to every question \citep{VQA2}. While VQA v2 resulted in a near 50/50 balance for Yes/No questions, the distribution for non-binary questions (e.g., ``What type of...?''; ``What sport is...?'') remains skewed towards a handful of top answers \citep{VQA2}.

Given the difficulty of isolating bias from crowdsourced data, researchers have instead begun to emphasize generalization to new domains with different biases. In this line, \citet{VQA-CP} introduced VQA-CP, a re-division of the existing VQA datasets in which the distribution of answers per question type is inverted between train and test splits. For instance, in the VQA-CP v1 train split, ``tennis'' is the most frequent answer for the question ``What sport is...?'', while ``skiing'' is very uncommon; in the test split, this prior is reversed.
Most relevant to our work, \citet{ramakrishnan2018overcoming} applied \mbox{AdvReg} to VQA-CP, and found that it improved test performance over a non-regularized model.
Similarly, \citet{belinkov2019starsem} analyzed the effects of using \mbox{AdvReg} to address bias in NLI.
In this work, we analyze the effects of \mbox{AdvReg} on VQA models in further detail, complement AdvReg with a scheduling scheme, and point to remaining limitations in its behavior.

\pagebreak
% \vspace{-2pt}
\section{Methods}
%\vspace{-2pt}
\subsection{Adversarial Regularization}
%\vspace{-2pt}

Many modern VQA architectures adhere to a common modular design \citep{pythia18arxiv} consisting of the following four components:
% \pagebreak
\begin{itemize}[itemsep=2pt] %,topsep=3pt]
    \item $f_v(I; \theta_v): I \mapsto v$ Image encoder
    \item $f_q(Q; \theta_q): Q \mapsto q$ Question encoder
    \item $f_z(v,q; \theta_z): v,q \mapsto z$ Multimodal fusion
    \item $g_{\text{VQA}}(z; \theta_\text{VQA}): z \mapsto P(a)$ Answer classifier
\end{itemize}
Composing these components, we obtain the following expression for the base VQA model. This model is trained to minimize cross entropy loss:\footnote{Since the VQA evaluation metric includes ground truth answers from 10 different subjects, we follow the top-performing models in using a soft target, multi-label variant of the cross entropy objective (see \citealt{teney2018tips}).}
%
% \vspace{-2pt}
\begin{align}
    P(a | I, Q) &= g_{\text{VQA}}(f_z(f_v(I), f_q(Q)))\\
    \mathcal{L}_{\text{VQA}} &= - \sum_{i} a_i \log P(a_i | I, Q)
\end{align}
\vspace{-12pt}

In \mbox{AdvReg}, we introduce an adversarial classifier $g_{\text{ADV}}(q; \theta_\text{ADV})$, which attempts to infer the correct answer from only the question features. $g_{\text{ADV}}$ shares the same question feature extractor $f_q$ as the base VQA model. However, $f_q$ and $g_{\text{ADV}}$ are separated by a gradient reversal layer (GRL). The GRL is a pseudo-function that negates gradients on the backward pass; otherwise, it leaves inputs unchanged:
\vspace{-2pt}
\begin{equation}
    \text{GRL}_{\lambda}(x) = x \label{eq:grl} \hspace{2em}
    \frac{\partial \text{GRL}_{\lambda}}{\partial x} = -\lambda_{\text{GRL}}
\end{equation}
%\vspace{-2pt}
where $\lambda_{\text{GRL}}$ is a hyperparameter. As above, the adversary is trained to minimize the cross entropy loss $\mathcal{L_{\text{ADV}}}$:
\vspace{-2pt}
\begin{align}
    P(a | Q) &= g_{\text{ADV}}({\text{GRL}}_{\lambda}(f_q(Q)))\\
    \mathcal{L}_{\text{ADV}} &= - \sum_{i} a_i \log P(a_i | Q)
\end{align}

\vspace{-6pt}
The adversarial relationship between the main model and the adversary can be expressed as:
\vspace{-2pt}
\begin{equation}
    \min_{\theta_{v, q, z, \text{VQA}}} \hspace{0.5em} \max_{\theta_{q, \text{ADV}}} \mathcal{L} = \mathcal{L_{\text{VQA}}} - \lambda_{\text{ADV}}\mathcal{L_{\text{ADV}}}
\end{equation}
where the regularization coefficient $\lambda_{\text{ADV}} \geq 0$ controls the trade-off between performance on VQA and robustness to language bias. Additionally, $\lambda_\text{GRL} \geq 0$ (from Eq. \ref{eq:grl}) scales the reversed gradients. These two hyperparameters perform related, but different, functions. Setting either or both to zero disables the regularization, since $f_q$ receives no gradients from the adversary. This combination is equivalent to the baseline model. Meanwhile, setting $\lambda_\text{ADV} > 0, \lambda_\text{GRL} > 0$ enables \mbox{AdvReg}. This setting is the main focus of our experiments.

\begin{figure}[t]
\includegraphics[width=\columnwidth]{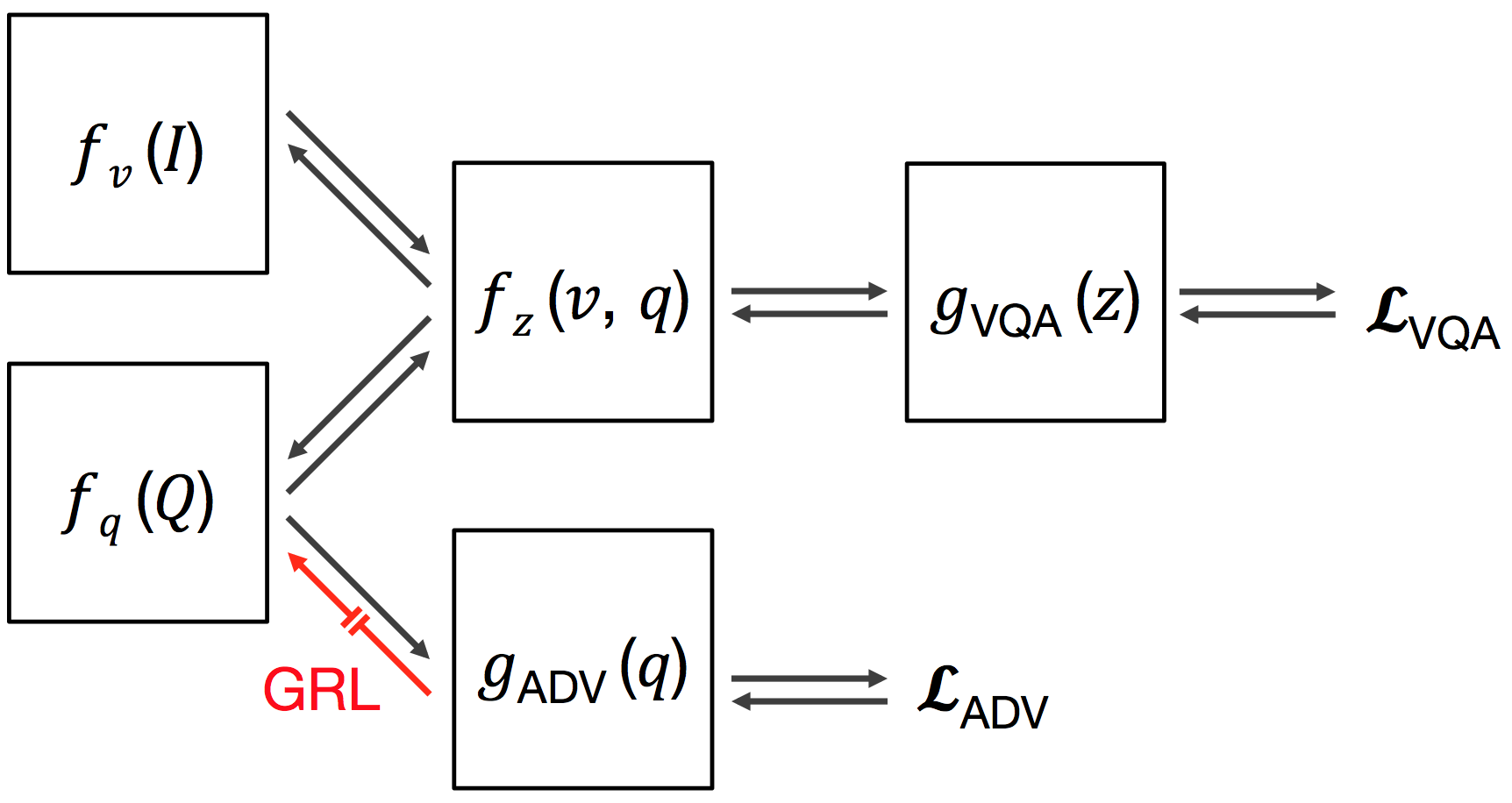}
\caption{Schematic diagram of adversarial VQA architecture. Right and left arrows represent forward and backward propagation, respectively. The red arrow indicates the gradient reversal layer.}
\label{fig:adversarial_schematic}
\vspace{-10pt}
\end{figure}

\subsection{Gradient Reversal Layer Scheduling}\label{sec:grl_scheduling}

Because the GRL counteracts the main gradient updates, \mbox{AdvReg} produces noisy gradients that can interfere with learning, as we observe in the experiments below (\cref{fig:grl_schedule}). To improve stability during the early stages of training, we experiment with a scheduling regime for the gradient reversal layer similar to that used in domain-adversarial neural networks \citep{ganin2016domain}. During training, we delay the introduction of regularization for the first $\mu$ iterations, which allows $f_q$ to receive clean gradients from the VQA model. Next, we have a warmup phase for $w$ iterations, in which we increase $\lambda_{\text{GRL}}$ linearly from $0$ to some constant $c$:
\vspace{-2pt}
\begin{equation}
\lambda_{\text{GRL}}(t) = \begin{cases} 
      0 & t\leq \mu \\
      \frac{c(t - \mu)}{w} & \mu\leq t\leq \mu + w \\
      c & t > \mu + w
   \end{cases}
\end{equation}
GRL scheduling introduces two new hyperparameters, $\mu$ and $w$, which we set by grid search; further details are given in \cref{app:grl}.

%\vspace{-2pt}
\section{Experimental Setup}
%\vspace{-2pt}
\subsection{Data}

\begin{table*}[ht!]
\centering
\small
\setlength{\tabcolsep}{4pt}
\begin{tabular}{l@{\hskip 2\tabcolsep}cc@{\hskip 2\tabcolsep}cccc@{\hskip 4\tabcolsep}cccc@{\hskip 2\tabcolsep}c}
\toprule
 &  &  & \multicolumn{4}{c}{VQA-CP v1 (test)} & \multicolumn{4}{c}{VQA-CP v1 (val)} & \multicolumn{1}{c}{VQA v1 (val)} \\
 \cmidrule(r{2.2em}){4-7} \cmidrule(r){8-11}
 Model & $\lambda_{\text{ADV}}$ & $\lambda_{\text{GRL}}$ & Overall & Yes/No & Num. & Other & Overall & Yes/No & Num. & Other & Overall \\ \midrule
Baseline & 0 & 0 & 37.87 & 42.58 & 14.16 & 42.71 & \textbf{65.79} & 86.98 & 40.06 & 56.41 & \textbf{62.68} \\
+ \mbox{AdvReg} & 0.01 & 0.1 & \textbf{45.69} & 77.64 & 13.21 & 26.97 & 46.94 & 65.32 & 32.95 & 37.22 & 46.34 \\
+ GRL Sch. & 0.01 & 0.1 & 44.09 & 75.01 & 13.40 & 25.67 & 46.45 & 67.28 & 29.11 & 35.71 & 46.71 \\ \midrule
 &  &  & \multicolumn{4}{c}{VQA-CP v2 (test)} & \multicolumn{4}{c}{VQA-CP v2 (val)} & VQA v2 (val) \\
  \cmidrule(r{2.2em}){4-7} \cmidrule(r){8-11} 
Baseline & 0 & 0 & 38.80 & 41.70 & 12.17 & 44.59 & \textbf{67.76} & 84.76 & 49.22 & 57.04 & \textbf{63.27} \\
+ \mbox{AdvReg} & 0.005 & 1 & 36.33 & 59.33 & 14.01 & 30.41 & 50.63 & 67.39 & 38.81 & 38.37 & 48.78 \\
+ GRL Sch. & 0.005 & 1 & \textbf{42.33} & 59.74 & 14.78 & 40.76 & 56.90 & 69.23 & 42.50 & 49.36 & 51.92 \\ \bottomrule
\end{tabular}
\caption{Performance comparison of baseline and adversarially-trained models on VQA-CP/VQA v1 and v2 datasets using the best-performing hyperparameters. %Adversarial regularization markedly increases performance on VQA-CP test, indicating improved generalization to out-of-domain examples. However, these gains come at the cost of substantially reduced performance on in-domain data on the VQA-CP and VQA validation sets.
}
\label{table:vqa_cp_results}
\vspace{-10pt}
\end{table*}

We evaluated the performance of our \mbox{AdvReg} setup on VQA-CP v1 and v2 \citep{VQA-CP}. We also retrained our best-performing models with the same hyperparameter settings on VQA v1~\citep{VQA1} and v2~\citep{VQA2} in order to evaluate performance on datasets without changing priors.

One difficulty of working with VQA-CP is the lack of validation sets. \citet{ramakrishnan2018overcoming} explain that VQA-CP does not provide validation sets due to the difficulty in varying the answer distributions of binary questions across more than two splits. The authors note that, in place of early stopping, they train their models ``until convergence.''\footnote{In correspondence, the authors clarified that they trained for a fixed interval determined by the number of iterations to reach peak performance on VQA v2. Since overfitting tends to occur more rapidly on VQA-CP, we view an in-domain val split as a more reliable early stopping metric.} Although the nonstandard structure of VQA-CP makes validation tricky, we believe it is important to have some mechanism to distinguish between overfitting to language priors and overfitting to the examples in the training set (the latter may occur regardless of the presence of language biases). Our solution is to train models on 90\% of the training data and reserve the remaining 10\% (sampled randomly) for validation. Score on the val split is useful as an early stopping metric, but does not forecast test performance. In this way, we are able to prevent our models from overfitting to the training data, while remaining agnostic to the distribution of priors in the test set.

\begin{figure*}[ht!]
\includegraphics[width=\textwidth]{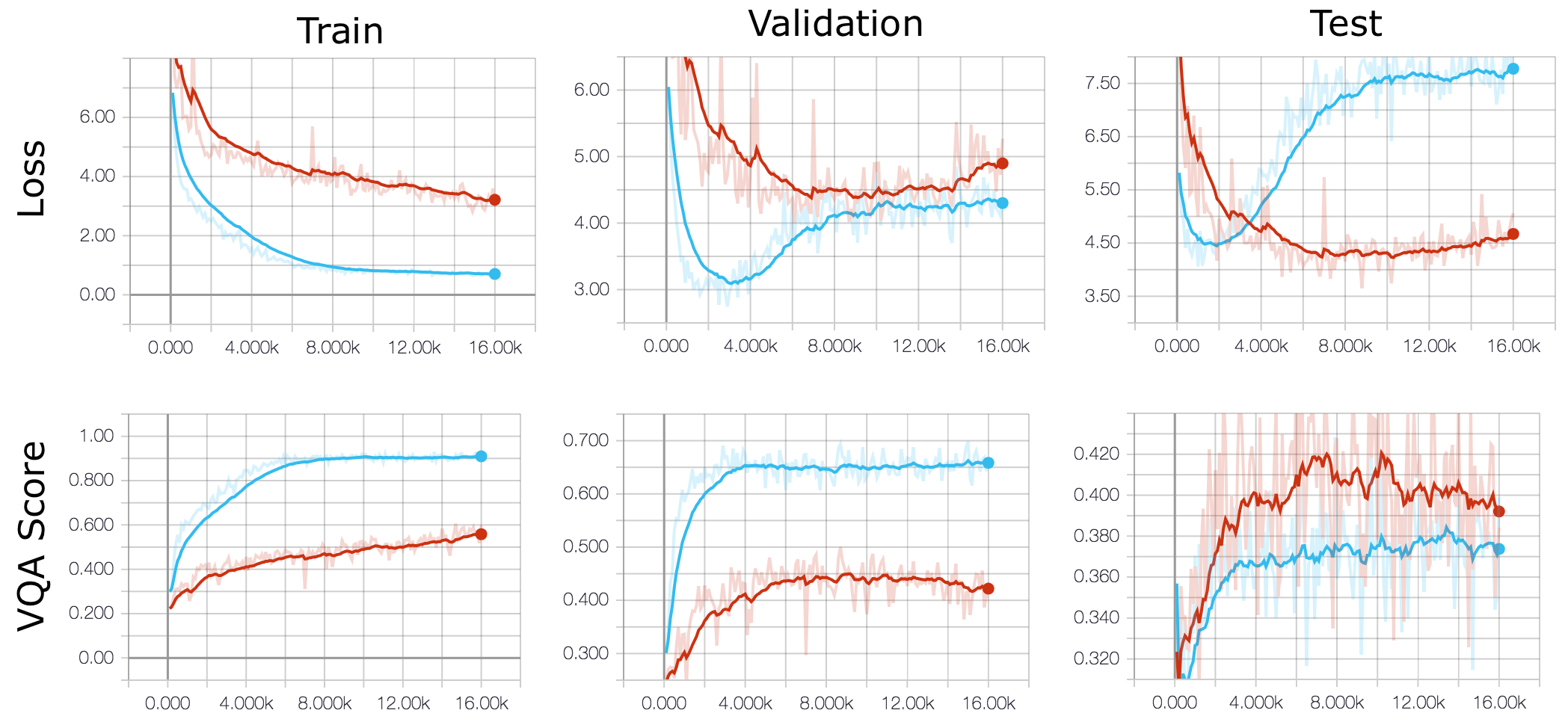}
\caption[VQAv1-CP]{Comparison of regularized (red) and baseline (blue) models on VQA-CP v1 train, val, and test. The baseline model exhibits severe overfitting on both the val and test splits. In contrast, the regularized model overfits much less and achieves a higher score on VQA-CP test.}
\label{fig:trainvaltest}
\vspace{-10pt}
\end{figure*}

While the addition of a VQA-CP val set enables early stopping, models that perform best on the val set will tend to be under-regularized, since \mbox{AdvReg} reduces in-domain performance. We considered creating a second val set derived from VQA-CP test for model selection. However, in addition to introducing additional complexity, this approach would both compromise our ability to remain agnostic to the test set and make our results incomparable with prior work. Therefore, we follow \citet{ramakrishnan2018overcoming} and perform model selection on VQA-CP test. However, to increase transparency, we report results across a broad range of hyperparameters. We hope that recognition of these challenges will motivate the introduction of a standard val set for VQA-CP.

%\vspace{-2pt}
\subsection{Implementation}

Our experimental setup is based on the Pythia implementation of the Bottom-Up / Top-Down VQA model \citep{pythia18software, anderson2018bottom}.\footnote{Our code is available at \url{https://github.com/gabegrand/adversarial-vqa}} The adversarial classifier $g_{\text{ADV}}$ is implemented as a two-layer fully-connected network with 512 hidden units and ReLU activation. Unless otherwise noted, we use the default hyperparameters from Pythia. Additional details are available in \cref{app:implementation}. 
\section{Results}

\subsection{Strengths of \mbox{AdvReg}}

\cref{table:vqa_cp_results} summarizes the results of the baseline model and the best performing adversarially regularized models. 
On the VQA-CP v1 test set, our best \mbox{AdvReg} model outperforms the baseline by 7.82\%, attaining a new state-of-the-art for this task.
On the VQA-CP v2 test set, our best \mbox{AdvReg} model performs worse than the baseline; however, with GRL scheduling, it surpasses the baseline by 3.53\%, again setting a new state-of-the-art.
Note that in both cases, our models perform better than \citet{ramakrishnan2018overcoming}, who report scores of 43.43\% and 41.17\% on VQA-CP v1 and v2 test, despite the fact that we use only 90\% of the available training data. This result indicates that allocating 10\% for validation helps prevent overfitting to the training examples.

To highlight how \mbox{AdvReg} mitigates overfitting, \cref{fig:trainvaltest} plots loss curves of the baseline (blue) and regularized (red) models during training. The baseline model exhibits severe overfitting on both VQA-CP v1 val and test. Note that overfitting on the test set appears around 2000 iterations as the model begins to over-rely on language priors. In contrast, overfitting on the val set appears later (around 3500 iterations) as the model begins to memorize the training examples.

\begin{figure*}[t]
\includegraphics[width=\textwidth]{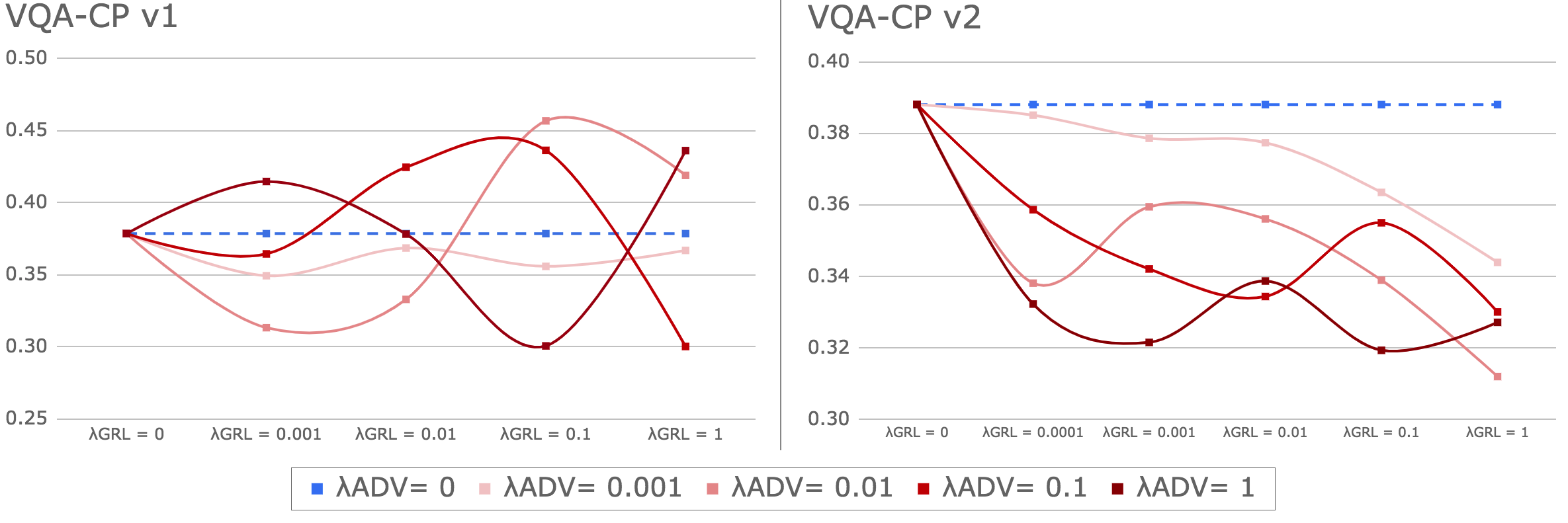}
\vspace{-20pt}
\caption{Hyperparameter sweep on VQA-CP v1 and v2 test. Each line is a different setting of $\lambda_{\text{ADV}}$; lighter/darker red indicates less/more regularization, respectively. $\lambda_{\text{GRL}}$ is varied along the x-axis. Blue dashes: baseline score.}
\label{fig:tuning}
% \vspace{-10pt}
\end{figure*}

In general, \mbox{AdvReg} works well out-of-box on VQA-CP v1. Many of the hyperparameter combinations we tested (\cref{fig:tuning}) outperform the baseline on VQA-CP v1 test. The key to successful regularization appears to be balancing $\lambda_{\text{ADV}}$ and $\lambda_{\text{GRL}}$. As \cref{fig:tuning} reveals, large values of $\lambda_{\text{ADV}}$ perform better with small values of $\lambda_{\text{GRL}}$, and vice-versa. However, when $\lambda_{\text{ADV}}$ is too small, \mbox{AdvReg} fails to improve performance; %yield any performance improvements; 
none of the models we tested with $\lambda_{\text{ADV}} = 0.001$ outperformed the baseline. On the other hand, when $\lambda_{\text{ADV}}$ is too large, training becomes unstable; for $\lambda_{\text{ADV}} > 1$ (not shown), we observed many training runs failing %that many training runs failed 
to converge due to exploding gradient values.

\subsection{Shortcomings of \mbox{AdvReg}}

The improved performance on the out-of-domain test sets comes at the expense of performance on the in-domain validation sets. As \cref{table:vqa_cp_results} shows, on both VQA-CP v1 and v2 val, \mbox{AdvReg} models significantly under-performed the baseline (\mbox{-18.85\%} and \mbox{-10.66\%}, respectively). Retraining with the same hyperparameters on the original VQA v1 and v2 datasets yielded similar results.

Notably, these findings differ from \citet{ramakrishnan2018overcoming}, who report only minimal reductions in performance on VQA v1 and v2 from \mbox{AdvReg}. One explanation is that the gains we observed on VQA-CP test relative to \citeauthor{ramakrishnan2018overcoming} resulted in diminished performance on VQA-CP val. Indeed, across all runs of our experiments, we found that score on VQA-CP v1 test correlated negatively with score on the val split ($r^2$ = -0.355, $p$ = 0.013).\footnote{We did not find a significant correlation between test and val performance on VQA-CP v2 ($r^2$ = 0.237, $p$ = 0.141).} In their work, \citeauthor{ramakrishnan2018overcoming} also introduce a secondary ``difference of entropies'' (DoE) regularizer, which they find improves in-domain performance and helps to stabilize adversarial training. However, even without DoE, they report margins of only 1-4\% between their \mbox{AdvReg} and baseline models. 
Ultimately, these unaccounted differences may be due to implementation details, suggesting the need for a closer comparison.\footnote{To our knowledge, code from \cite{ramakrishnan2018overcoming} is not public at present.}

Our results also highlight interesting differences between VQA-CP v1 and v2. On VQA-CP test, the gains due to \mbox{AdvReg} were more significant on v1 as compared to v2. However, on the validation sets, the losses were also greater. This pattern also applied with respect to the original versions of these datasets (i.e., VQA v1 and v2). These findings support the notion that VQA v2 is indeed less biased than v1.

%\vspace{-2pt}
\subsection{Effect of GRL scheduling}

Without GRL scheduling, none of the \mbox{AdvReg} hyperparameter combinations we tested outperformed the baseline on VQA-CP v2 test (see \cref{fig:tuning}). This finding may be attributed to the substantial amount of noise that the adversary injects into the gradient updates for the question encoder, as demonstrated by recording gradient norms throughout training.
 
\begin{figure}[t]
\centering
\includegraphics[width=\columnwidth]{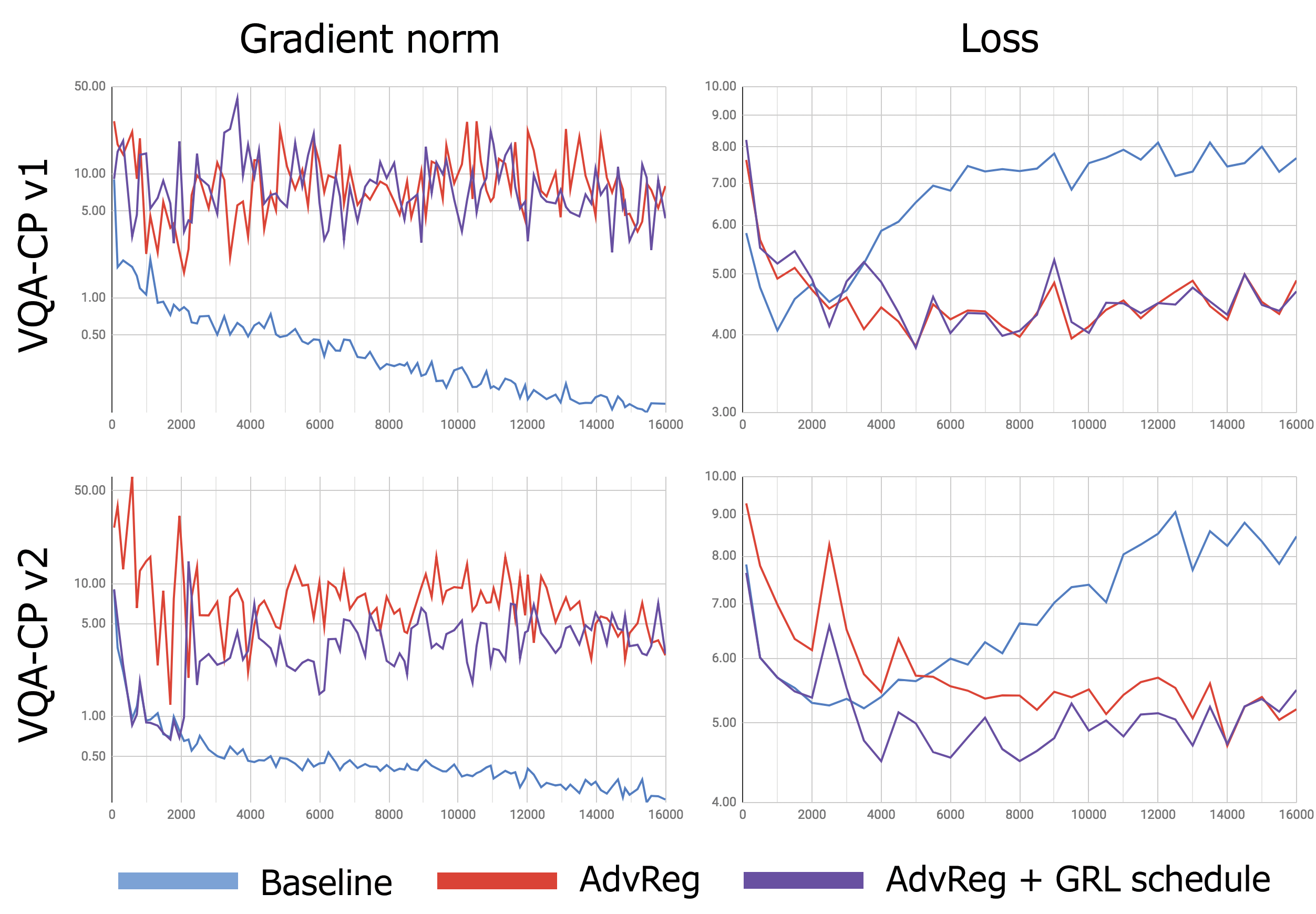}
\caption{Gradient norms and loss during adversarial training. On VQA-CP v2, GRL scheduling helps to reduce gradient noise early in training (bottom left), leading to lower loss values (bottom right). %Meanwhile, 
On VQA-CP v1, the baseline %model 
(blue, top right) overfits more quickly; hence, delaying the regularization is less effective.}
\label{fig:grl_schedule}
\vspace{-10pt}
\end{figure}

As \cref{fig:grl_schedule} illustrates, on VQA-CP v2, GRL scheduling reduces gradient instability early in training, allowing the model to converge to a lower loss value. In the best-performing schedule, regularization was delayed until $\mu$ = 2000 iterations, and slowly warmed up for the following $w$ = 4000 steps. This schedule resulted in a 6.00\% performance increase on VQA-CP v2 test compared to using the same regularization coefficients without GRL scheduling, and a 3.53\% improvement over the baseline (see \cref{table:vqa_cp_results}).

On VQA-CP v1, we did not observe commensurate improvements from GRL scheduling. We hypothesize that introducing \mbox{AdvReg} on a delay may not be as effective on v1 due to the more prominent biases in this dataset. Note that the baseline model begins to overfit roughly twice as quickly on VQA-CP v1 as on VQA-CP v2 (\cref{fig:grl_schedule}, Baseline loss). Accordingly, in addition to sweeping the same hyperparameters tested on VQA-CP v2, we experimented with accelerated GRL schedules for VQA-CP v1. While five of the runs outperformed the baseline, three of these were with no start delay. Moreover, all of the runs with GRL scheduling performed worse than a model with the same regularization coefficients with static $\lambda_{\text{GRL}}$. Finally, many of the runs on VQA-CP v1, and especially those with fewer warm-up iterations, diverged due to exploding gradients. These findings suggest that the stronger the biases in a dataset, the earlier \mbox{AdvReg} must be introduced in order to counter overfitting effectively.
% The lack of success of GRL scheduling on VQA-CP v1 may thus be attributed to early overfitting, resulting from a larger discrepancy between train and test sets in v1; observe that \mbox{AdvReg} (without scheduling) is much more successful on v1 than on v2.

\begin{table*}[ht!]
\centering
\footnotesize
\setlength{\tabcolsep}{4pt}
\begin{tabular}{@{}lccccc|lccccc@{}}
\toprule
\multicolumn{6}{c}{\mbox{AdvReg} $>>$ Baseline} & \multicolumn{6}{c}{\mbox{AdvReg} $<<$ Baseline} \\ 
Question type & Ans. & N & Base. & Reg. & $\Delta$ & Question type & Ans. & N & Base. & Reg. & $\Delta$ \\\midrule
is there a & Yes/No & 6501 & 16.75 & 93.41 & 49.83 & is this & Yes/No & 13063 & 76.96 & 64.85 & -15.82 \\
is this a & Yes/No & 7177 & 29.70 & 86.27 & 40.60 & what color is the & Other & 4418 & 47.71 & 21.36 & -11.64 \\
are the & Yes/No & 5037 & 24.99 & 87.07 & 31.27 & what & Other & 8646 & 38.48 & 25.28 & -11.42 \\
does the & Yes/No & 3525 & 24.02 & 94.34 & 24.79 & what is the & Other & 6363 & 41.49 & 28.51 & -8.26 \\
is & Yes/No & 3154 & 32.84 & 92.38 & 18.78 & is the & Other & 1148 & 50.44 & 4.40 & -5.29 \\
are they & Yes/No & 1577 & 27.96 & 89.40 & 9.69 & what kind of & Other & 3141 & 51.43 & 35.51 & -5.00 \\
do you & Yes/No & 1083 & 26.14 & 92.32 & 7.17 & how many & Number & 15917 & 15.90 & 13.01 & -4.60 \\
is there & Yes/No & 5265 & 68.83 & 78.45 & 5.06 & what type of & Other & 1995 & 54.74 & 36.30 & -3.68 \\
is the person & Yes/No & 757 & 41.64 & 92.46 & 3.85 & none of the above & Other & 2057 & 29.65 & 13.66 & -3.29 \\
how many people are & Number & 2118 & 11.96 & 21.08 & 1.93 & what color are the & Other & 1435 & 56.93 & 35.74 & -3.04 \\ 
\bottomrule
\end{tabular}
\vspace{-4pt}
\caption{Comparison of relative strengths and weaknesses of regularized and baseline models. The top 10 question types for which the regularized model outperforms the baseline are shown on the left, and vice versa on the right.}
\label{table:quant_analysis}
\vspace{-10pt}
\end{table*}

\section{Error Analysis}
\vspace{-2pt}
We performed quantitative and qualitative error analyses to understand how \mbox{AdvReg} affects model inferences on different kinds of examples. To best highlight the effect of \mbox{AdvReg}, both analyses were performed on VQA-CP v1 test, where the change in priors is more pronounced. In both analyses, we compare our best \mbox{AdvReg} model (which did not use GRL scheduling) and the baseline model.

% \vspace{-2pt}
\subsection{Quantitative Analysis} \label{sec:analysis-quant}
%\vspace{-2pt}

We first explore how model performance differs by question type. In the VQA datasets, each question is assigned a type corresponding to the 64 most common prefixes (e.g., ``Is there a...?'') or ``none of the above.'' Additionally, each example is given an answer type (Yes/No, Number, Other).\footnote{Note that the mapping between question types and answer types is not exactly one-to-one. However, for a given question type, a single answer type typically predominates; therefore, we are able to draw an approximate correspondence between question and answer types.}

To quantify the relative performance of the \mbox{AdvReg} and baseline models, we computed a difference metric, weighted by the number of questions $N$ of the given type:
% \vspace{-2pt}
%
$$\Delta = \tfrac{N}{100} \big(\text{score}_\text{baseline} - \text{score}_\text{regularized}\big)$$
\cref{table:quant_analysis} shows the question types with the largest and smallest $\Delta$ values, respectively. Compared to the baseline, the \mbox{AdvReg} model excels at Yes/No examples, but suffers on Other examples. Overall, \mbox{AdvReg} improves Yes/No test performance by 35.06 points, but reduces Other performance by 15.74 points (\cref{table:vqa_cp_results}). Additionally, \mbox{AdvReg} reduces Number test performance by 0.95\%, though in general both models score poorly on counting questions---a known shortcoming of many VQA models \citep{chattopadhyay2017counting, trott2018interpretable, zhang2018learning}.

These results suggest that much of the observed advantage of \mbox{AdvReg} on VQA-CP test is due to the extreme biases present in the dataset. In VQA-CP, Yes/No questions encode very strong priors (e.g., ``no'' is the answer to roughly 90\% of the questions beginning with ``Is there a...?'' in the v1 training set). Because this prior is inverted, any learned association between question prefixes and answers becomes harmful at test time. That \mbox{AdvReg} scores well above chance (77.64\%) on Yes/No examples suggests that this model has, to a certain degree, learned to answer binary questions without relying on language priors.

In contrast, the 15.74\% drop on Other-type examples implies that \mbox{AdvReg} impairs the model's ability to make inferences about questions with heterogeneous answers. Other-type questions typically have 3--20 top answers. This finding suggests that \mbox{AdvReg} interferes with learning of language cues in the question that yield key information about the answer.

\subsection{Qualitative Analysis}

\begin{figure*}[pht!]
\centering
\begin{subfigure}[b]{0.48\textwidth}
\includegraphics[width=\linewidth]{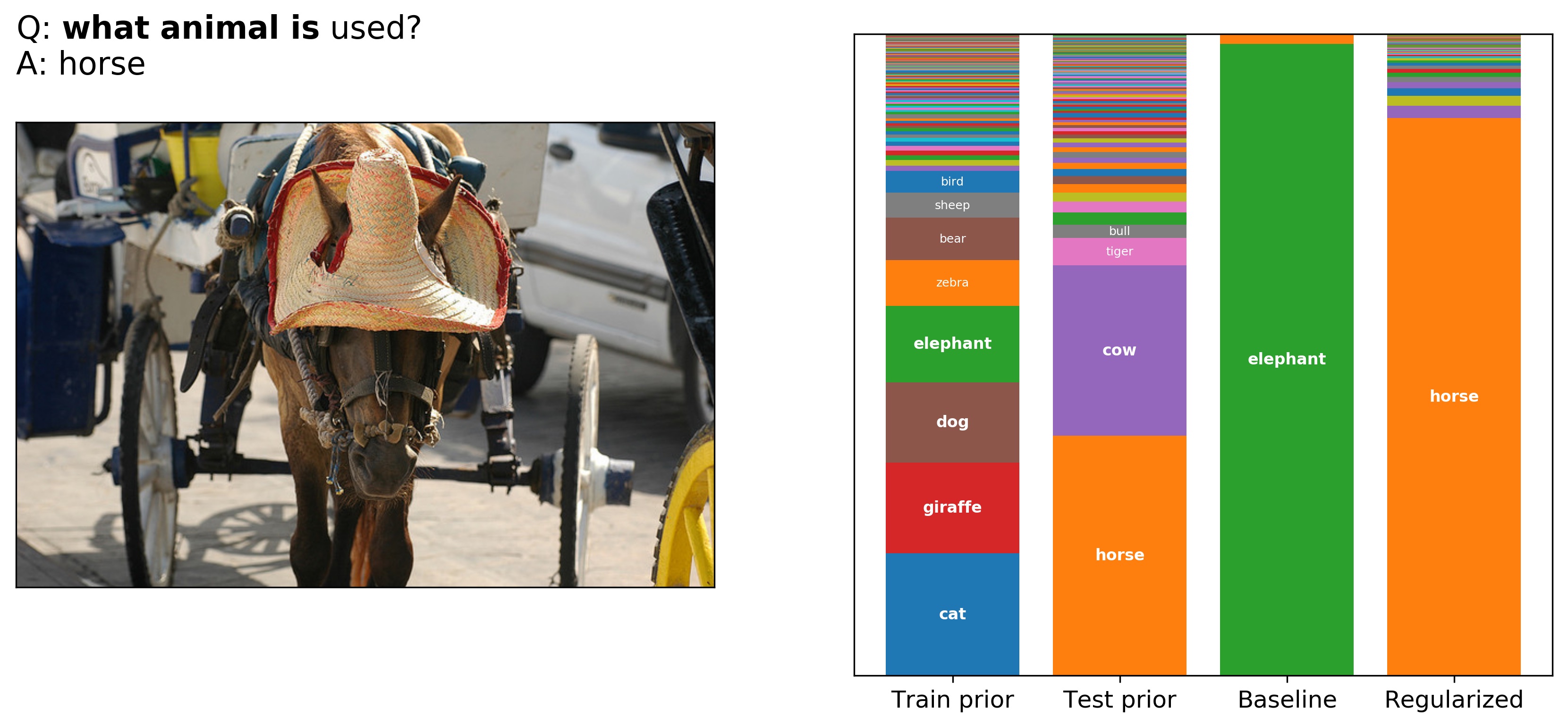}
%\caption{Regularized model fails to infer the correct form of the answer.}
\label{fig:qual-successes-horse}
\vspace{-8pt}
\end{subfigure}
\begin{subfigure}[b]{0.48\textwidth}
\includegraphics[width=\linewidth]{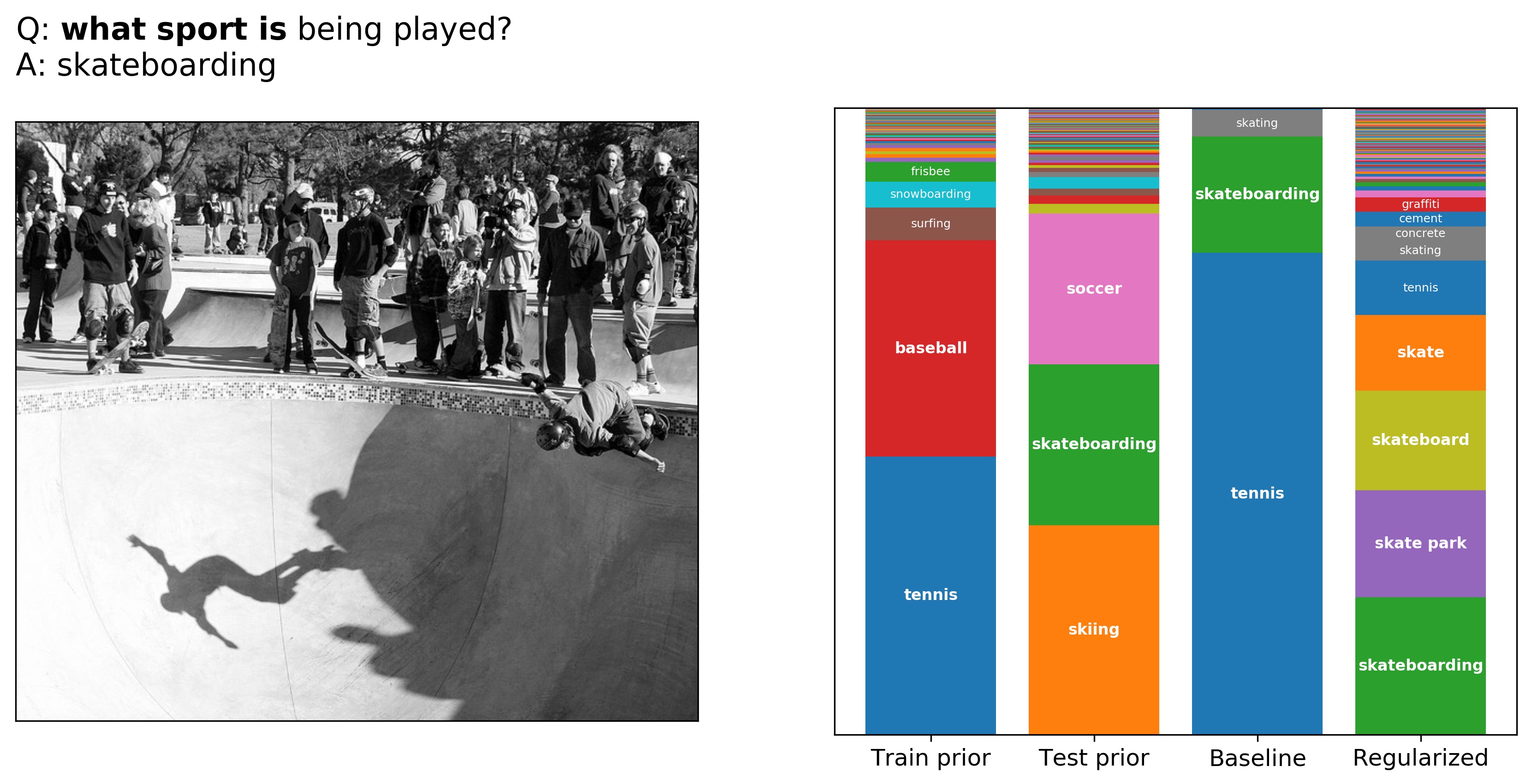}
%\caption{Regularized model fails to utilize real-world language priors.}
\label{fig:qual-successes-skateboarding}
\vspace{-8pt}
\end{subfigure} \\ 
\begin{subfigure}[b]{0.48\textwidth}
\includegraphics[width=\linewidth]{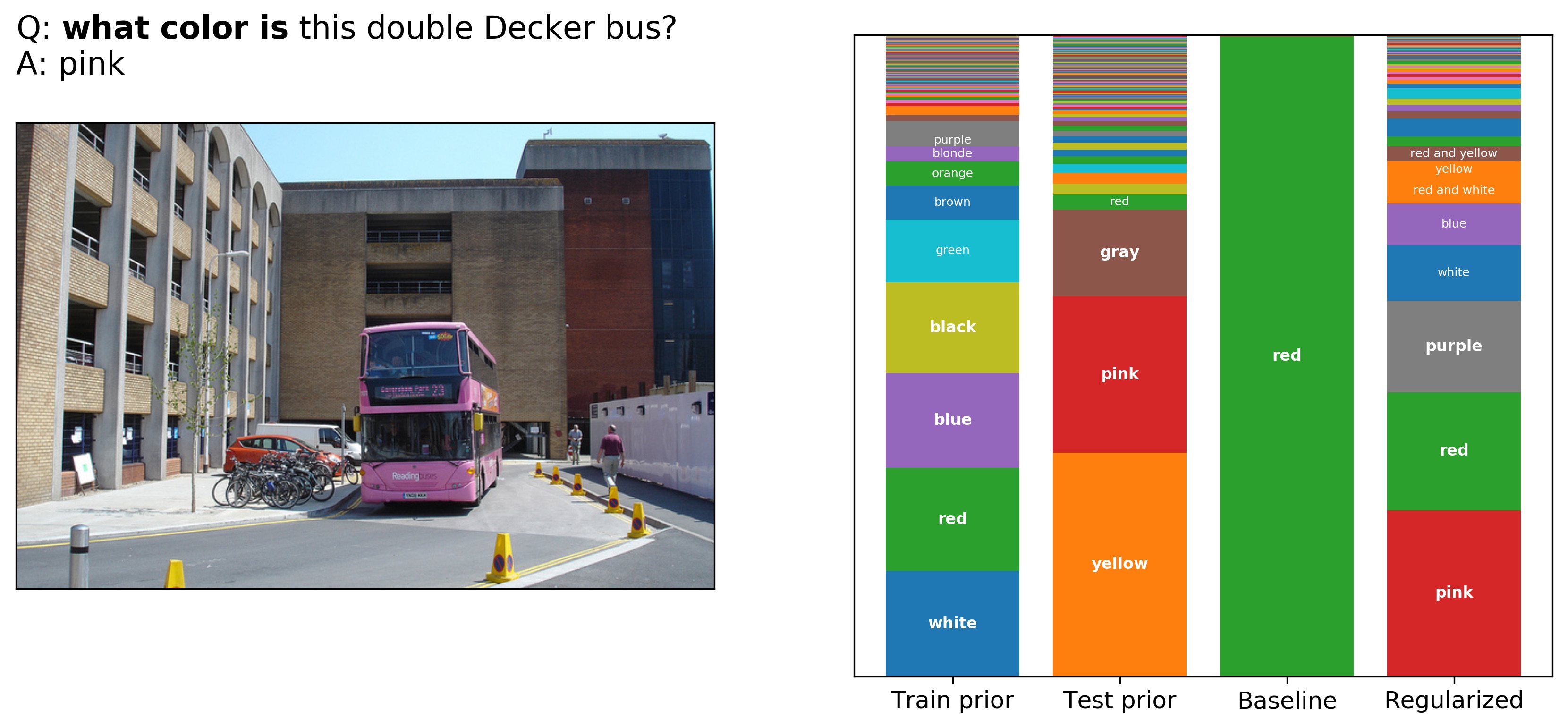}
%\caption{Regularized model distracted by visually-salient image features.}
\label{fig:qual-successes-pink}
\vspace{-5pt}
\end{subfigure}
\begin{subfigure}[b]{0.48\textwidth}
\includegraphics[width=\linewidth]{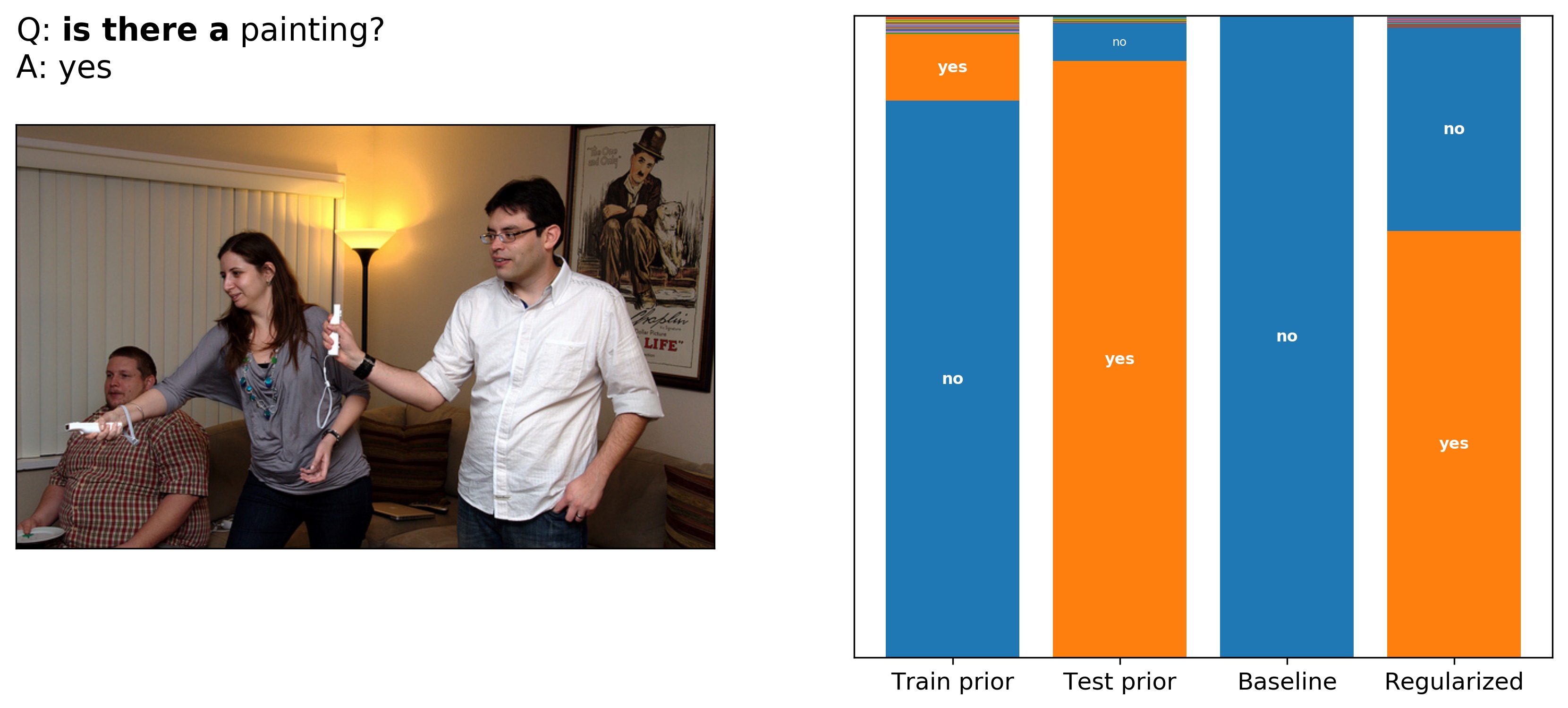}
%\caption{Regularized model relies on image features, while baseline model relies on language priors.}
\label{fig:qual-successes-painting}
\vspace{-5pt}
\end{subfigure}
\vspace{-6pt}
\caption{Visualization of \mbox{AdvReg} success cases. In each example, the leftmost two bars show the prior distribution over answers for the given question type (in bold). The rightmost two bars show the scores assigned to different answers by the baseline and \mbox{AdvReg} models for a particular example of the given type. The baseline model frequently assigns high probability to incorrect answers that are prominent in the training distribution. In contrast, the regularized model is able to make correct inferences in cases where the ground truth answer has low prior probability. Additional examples are provided in \cref{app:qualitative}.}
\label{fig:qualitative_successes}
\end{figure*}

In this section, we examine individual examples to highlight common success and failure modes of \mbox{AdvReg}. We consider different question types and compare the prior answer distribution in the train/test sets to the posterior distribution assigned by the \mbox{AdvReg} and baseline models. 
Expanding on the visualization format introduced by \citet[Fig.\ 3]{ramakrishnan2018overcoming}, %in prior work
%(\citealt{ramakrishnan2018overcoming}; Fig.\ 3), 
\cref{fig:qualitative_successes} shows examples where the \mbox{AdvReg} model successfully answered the question while the baseline model was wrong. 
In these cases, the baseline model prediction relies on the prior answer distribution in the train set, while the \mbox{AdvReg} model is able to overcome these priors to infer the correct answer.

% \pagebreak

\begin{figure*}[ph!]
\centering
\begin{subfigure}[b]{0.49\textwidth}
\includegraphics[width=\linewidth]{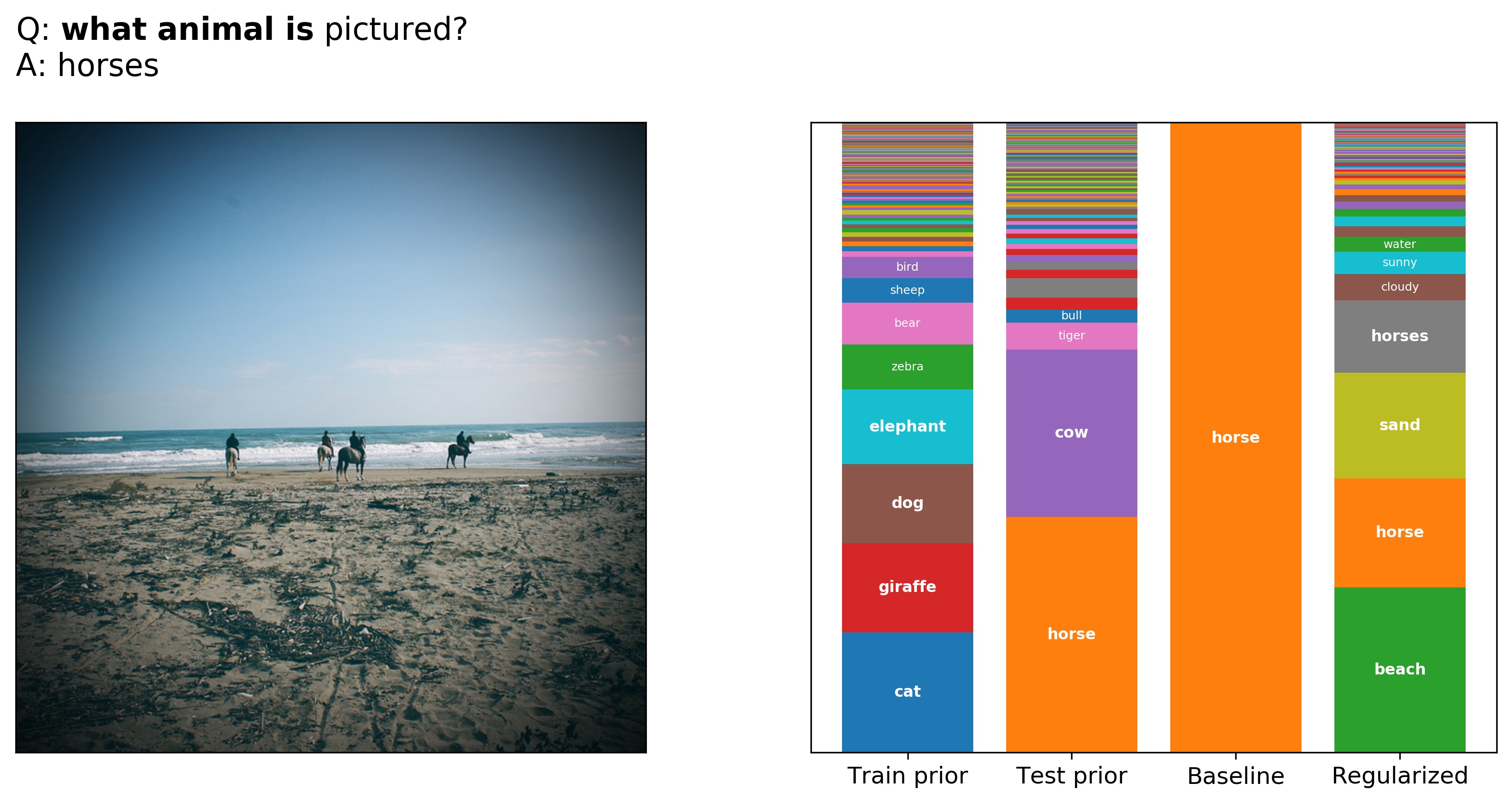}
\caption{\mbox{AdvReg} model fails to infer the correct form of the answer.}
\label{fig:qual-failures-horse}
\end{subfigure}
\begin{subfigure}[b]{0.49\textwidth}
\includegraphics[width=\linewidth]{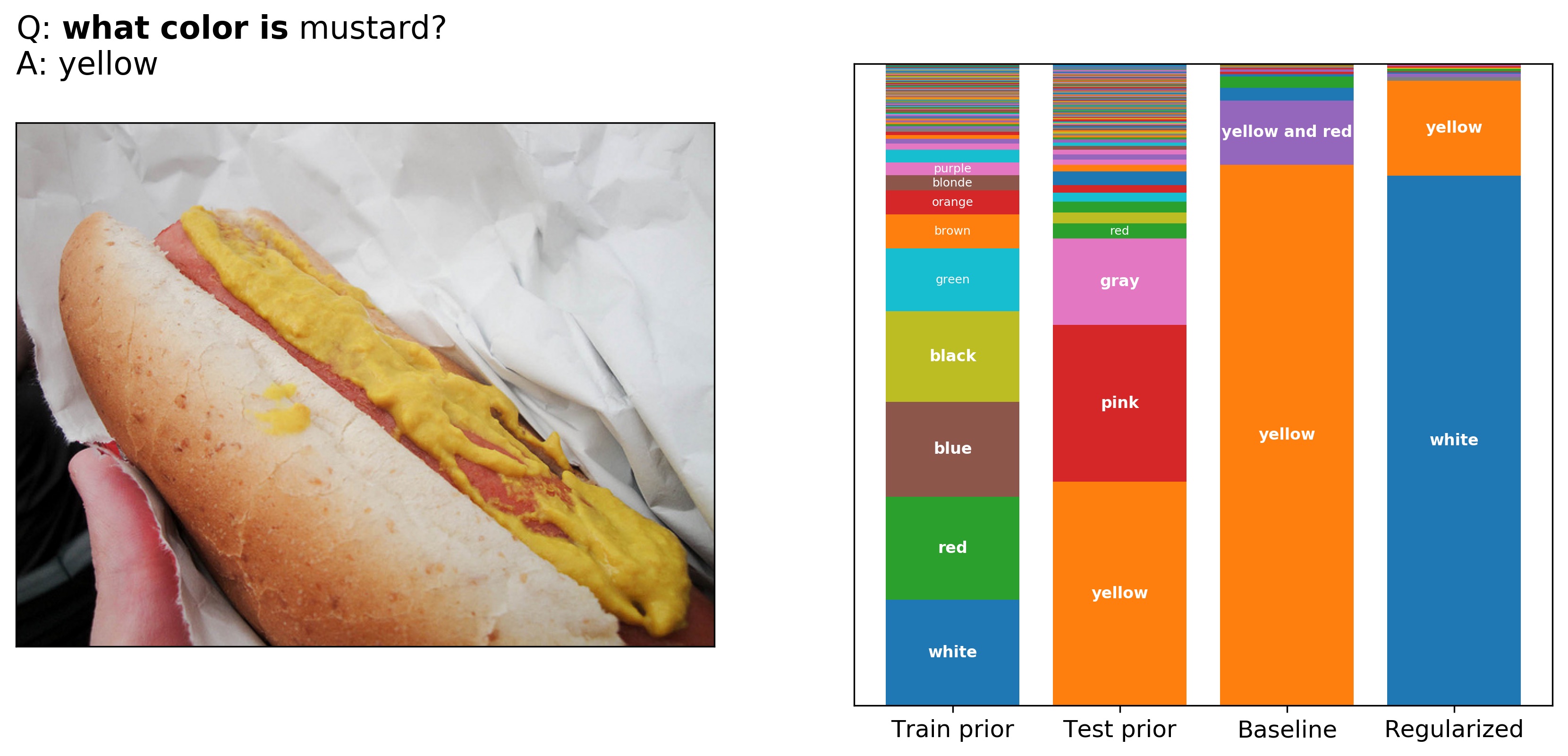}
\caption{\mbox{AdvReg} model fails to utilize real-world language priors.}
\label{fig:qual-failures-yellow}
\end{subfigure} \\ \vspace{12pt}
\begin{subfigure}[b]{0.49\textwidth}
\includegraphics[width=\linewidth]{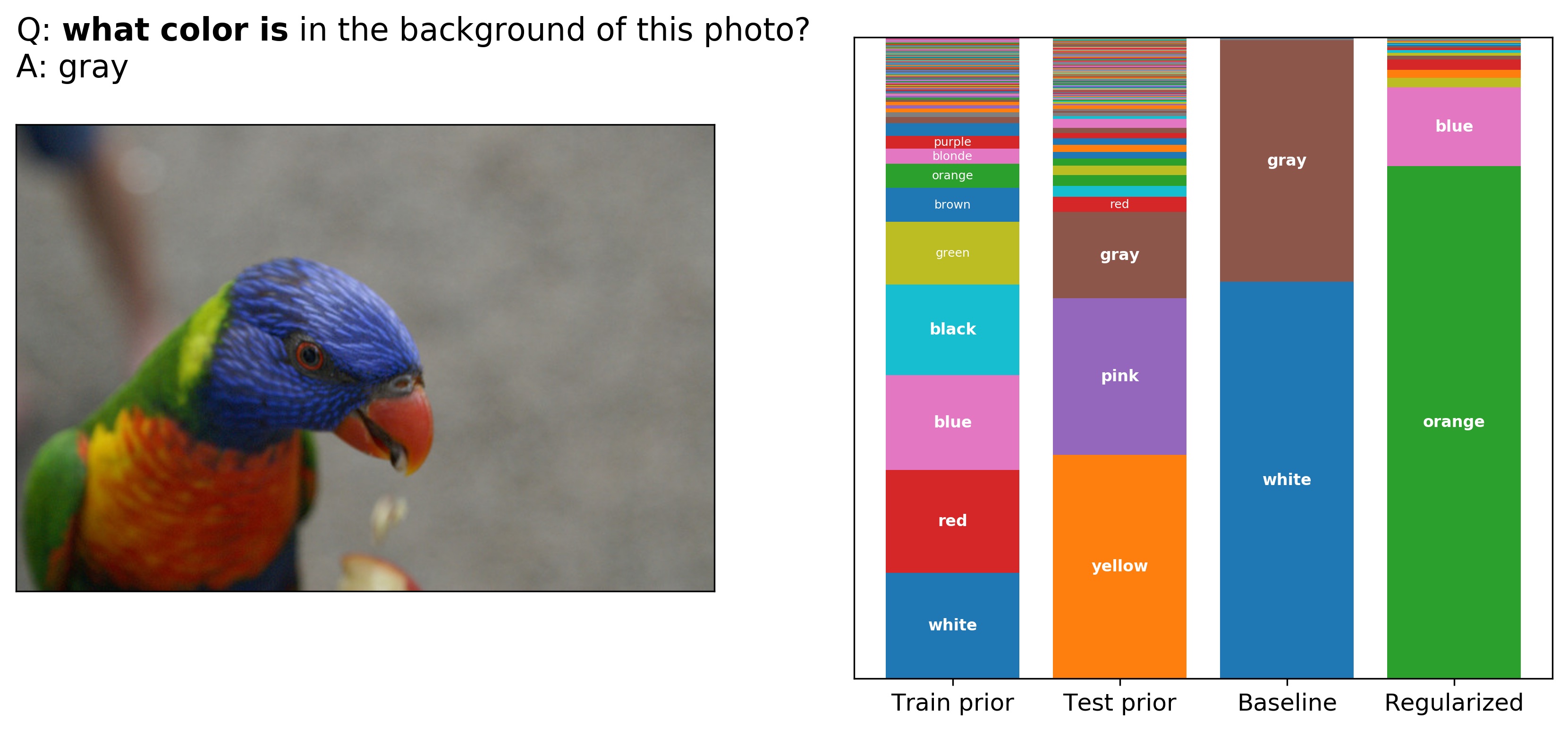}
\caption{\mbox{AdvReg} model distracted by visually-salient image features.}
\label{fig:qual-failures-gray}
\end{subfigure}
\begin{subfigure}[b]{0.49\textwidth}
\includegraphics[width=\linewidth]{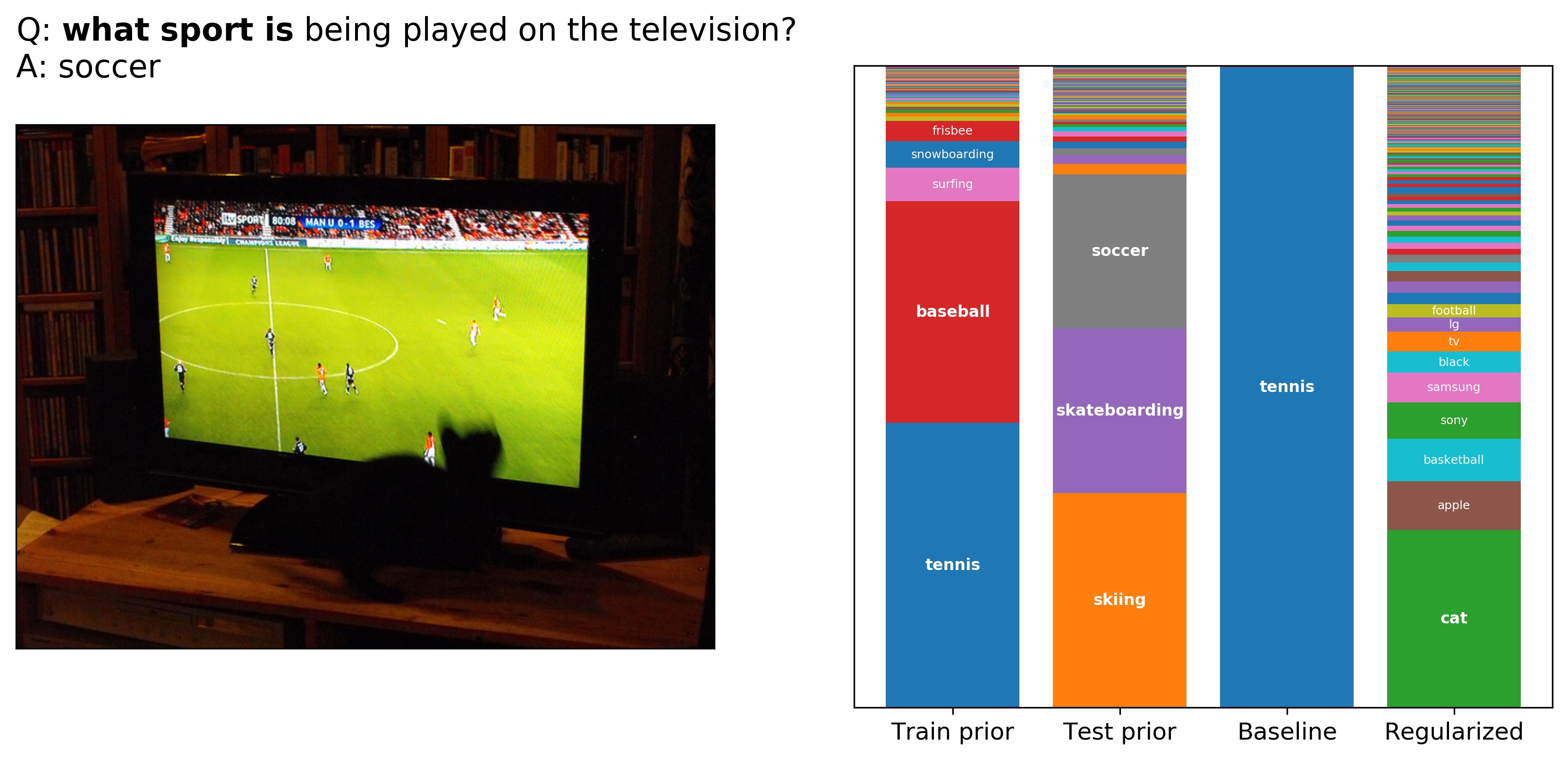}
\caption{\mbox{AdvReg} model relies on image features, while baseline model relies on language priors.}
\label{fig:qual-failures-soccer}
\end{subfigure}
\caption{Common failure modes of adversarial regularization. Additional examples are provided in \cref{app:qualitative}.}
\label{fig:qualitative_failures}
\end{figure*}

% \pagebreak

Turning to failures, we investigate what kinds of errors the \mbox{AdvReg} model makes on Other-type examples---the largest source of errors according to \cref{sec:analysis-quant}.
We randomly selected instances where the regularized model produced an incorrect answer, and manually grouped these examples into four approximate categories corresponding to different failure modes. 
%
%To further investigate what kinds of errors the regularized model makes on Other-type examples, we randomly selected examples from several question types with non-binary answers. For each example, we compared the answer probabilities produced by the baseline and \mbox{AdvReg} models against the VQA-CP v1 train and test priors. We collected several dozen instances where the regularized model produced an incorrect answer, and manually grouped these examples into four approximate categories corresponding to different failure modes. 
%
\cref{fig:qualitative_failures} shows representative examples for each of these failure modes; more examples are available in \cref{app:qualitative}. 

\cref{fig:qual-failures-horse} shows an example where the regularized model fails to infer the correct form of the answer from the question, answering ``beach'' to a question that entails animal answers. 
In \cref{fig:qual-failures-yellow}, the regularized model struggles with a question that relies on real-world language priors (i.e., mustard is yellow). 
In \cref{fig:qual-failures-gray}, the parrot's salient orange color distracts the regularized model from attending to the correct image region. 
\cref{fig:qual-failures-soccer} shows an example where  the regularized model relies on visual features %in the image 
(the cat), while the baseline %model 
relies on language priors (tennis is a common answer to sport questions). 
These findings suggest that \mbox{AdvReg} may encourage models to rely on visual features at the expense of learning to interpret task-relevant linguistic information.

%\vspace{-2pt}
\section{Conclusion}
%\vspace{-3pt}
% \parskip 0pt

In this work, we investigated several strengths and limitations of adversarial regularization, a recently introduced technique for reducing language biases in VQA models. Though we find \mbox{AdvReg} improves performance on out-of-domain examples in VQA-CP, one concern is that the pendulum has swung too far: there are both quantitative and qualitative signs that our models are over-regularized. Quantitatively, the performance of our \mbox{AdvReg} models suffers on in-domain examples in VQA-CP and the original VQA datasets. Additionally, while \mbox{AdvReg} boosts performance on binary questions, it impairs performance on other question types. Qualitatively, we observe that \mbox{AdvReg} models draw on salient image features while ignoring important linguistic cues in questions. These results demonstrate that \mbox{AdvReg} interferes with certain key aspects of reasoning.

Our findings highlight the need for further research in two areas: datasets and modeling. The lack of a validation set in VQA-CP makes it difficult to perform hyperparameter tuning in a principled way. Moreover, the exaggerated biases in the existing VQA-CP splits may encourage over-regularization, as evidenced by the sharp discrepancy between \mbox{AdvReg} performance on binary and non-binary question types. To address these issues, future iterations of VQA-CP could contain three or more splits with moderate but distinct ratios of Yes/No answers. Restructuring VQA-CP in this way would help balance the importance of binary and non-binary questions, while providing researchers with more sound evaluation metrics.

On the modeling side, our findings suggest that \mbox{AdvReg} requires further refinement to avoid impairing learning of task-relevant linguistic information. One possible approach would be to use attention to apply different amounts of regularization to different words in the question. In this way, regularization could be focused on the first few words of the question (e.g., ``Is there a...?'') that encode answer distribution biases, while preserving other useful linguistic information. Such enhancements could lead to more targeted regularization techniques that preserve the benefits of \mbox{AdvReg} while reducing the drawbacks discussed in this work.

\section*{Acknowledgements}
We would like to thank Alexander Rush for providing helpful advice and comments throughout our work on this project. 
GG and YB were supported by the Harvard Mind, Brain, and Behavior Initiative.

\bibliography{naaclhlt2019}
\bibliographystyle{acl_natbib}

% Added blank page here, as the appendix must be separate from the main submission.
\clearpage
\newpage

\onecolumn

\appendix
\section{Appendix}

\subsection{Implementation Details} \label{app:implementation}

Here, we provide additional details of our implementation. 
We experimented with different numbers of hidden layers $N = 1, 2, 3$ and hidden units $h = 256, 512, 1024, 2048$ in the adversarial classifier. We found the details of the adversary architecture to have little impact on performance, with the exception that adversaries with $N > 1$ hidden layers were more effective than one-layer adversaries.
Both the adversary and the base VQA model are randomly initialized with a fixed seed at the start of training. We co-train the networks for 16k iterations with two separate PyTorch Adamax optimizers with batch size 512 and learning rate 0.001. Unlike \citet{pythia18arxiv}, we keep the learning rate fixed throughout training to minimize the possibility of gradient scaling mismatch between the base model and the adversary. While this modification causes the performance of the baseline VQA model to drop 1.1\%, it greatly improves stability and convergence during adversarial training.

\subsection{GRL Scheduling Details}\label{app:grl}

For both VQA-CP v1 and v2, we performed a grid search to determine the optimal hyperparameters $\mu$ and $w$ for the GRL schedule. We tested all combinations of delay $\mu = 0, 1000, 2000, 3000, 4000, 5000, 6000$ and warmup duration $w = 1000, 2000, 3000, 4000$. Given that the baseline model demonstrates signs of overfitting on VQA-CP v1 as early as 2000 iterations into training, we tested an additional set of accelerated GRL schedules for VQA-CP v1 that consisted of all combinations of $\mu = 500, 1000, 1500, 2000, 2500, 3000, 3500$ and $w = 500, 1000, 2000, 4000$. 

Sometimes when \mbox{AdvReg} is introduced on a delayed schedule (especially if the value of $\mu$ is large), overfitting occurs before \mbox{AdvReg} takes effect. To avoid ending training prematurely, we always train for at least $\mu$ iterations before early stopping can be triggered. For instance, if $\mu = 3000$, then the earliest that we will stop training is $t = 4000$. For the purposes of evaluation, we also consider only scores from $t > \mu$ when scoring models under GRL scheduling.

\clearpage
\pagebreak

\subsection{Additional Examples} \label{app:qualitative}

% Somewhat janky but it gets the job done
\begin{figure}[h!]
\onecolumn
\includegraphics[width=\textwidth]{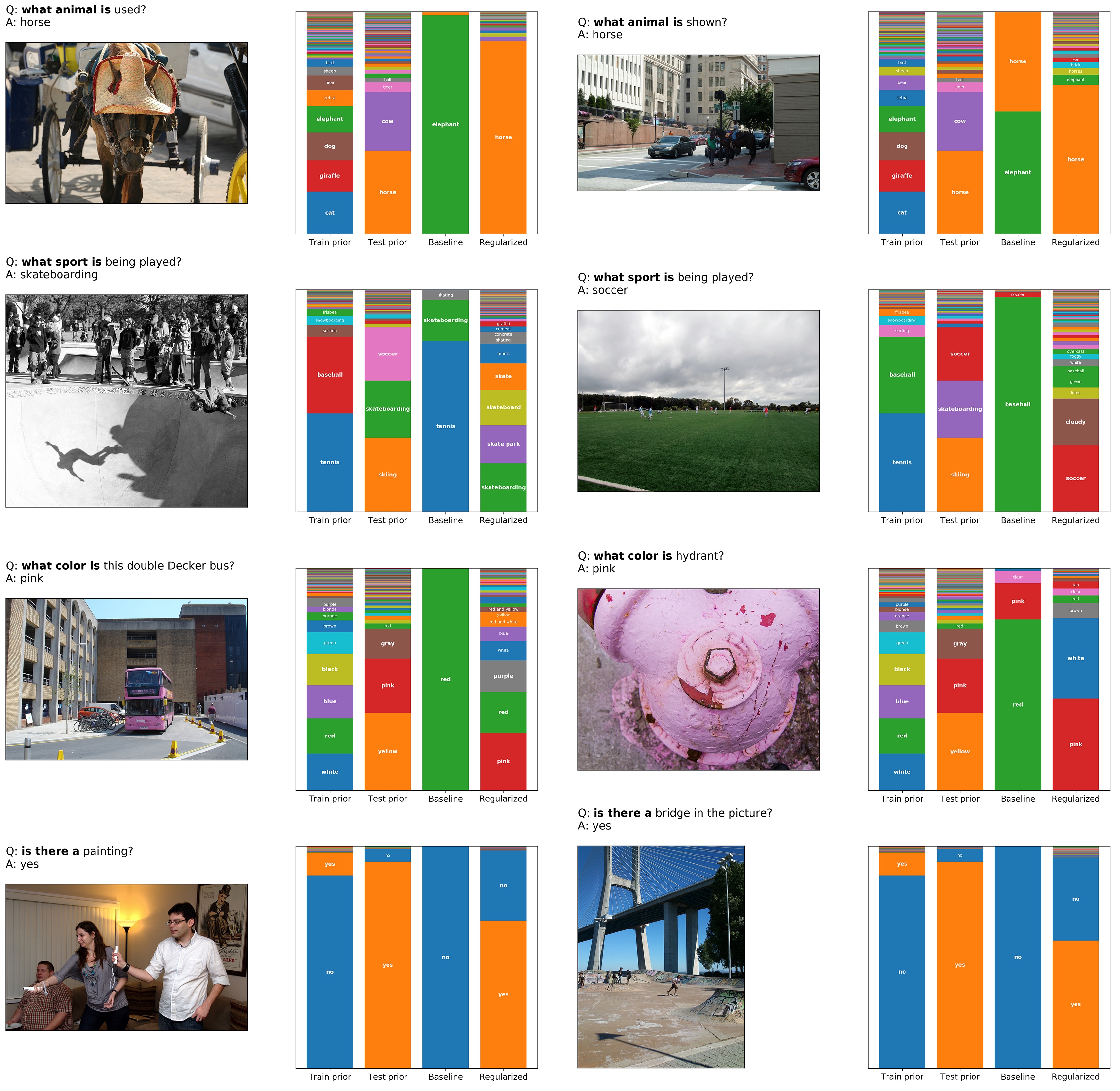}

\captionsetup{width=\textwidth}

\caption{Visualization of \mbox{AdvReg} success cases. In each example, the leftmost two bars show the prior distribution over answers for the given question type (in bold). The rightmost two bars show the scores assigned to different answers by the baseline and \mbox{AdvReg} models for a particular example of the given type. The baseline model frequently assigns high probability to incorrect answers that are prominent in the training distribution. In contrast, the regularized model is able to make correct inferences in cases where the ground truth answer has low prior probability.}
\label{fig:app:qualitative_success}
\end{figure}

\begin{figure*}[t]
\includegraphics[width=\textwidth]{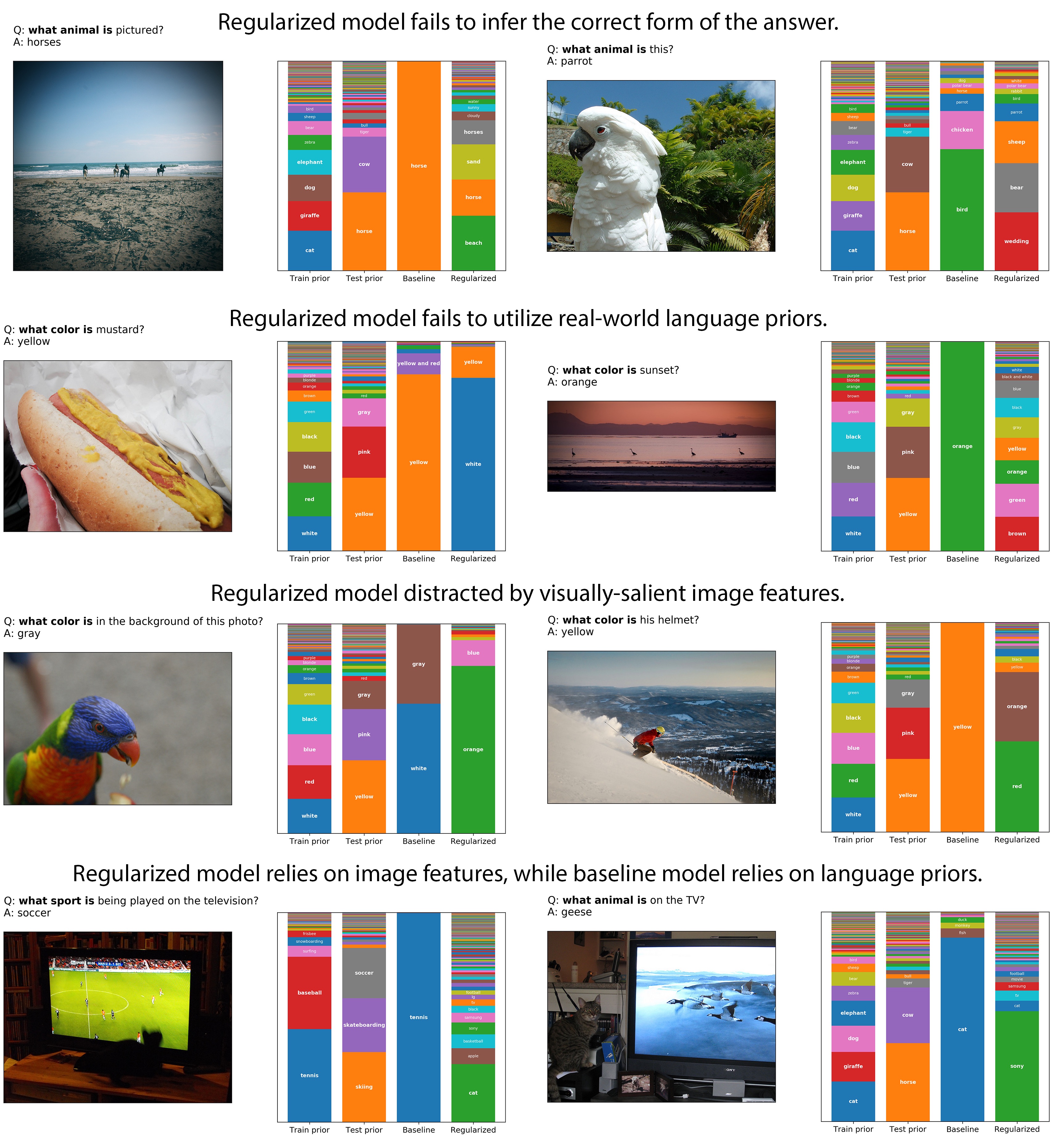}
\caption{Common failure modes of adversarial regularization. First row: the regularized model fails to infer the correct form of the answer from the question, answering ``beach'' and ``wedding'' to questions that entail animal answers. Second row: the regularized model struggles with questions that rely on real-world language priors; i.e., mustard is yellow, sunset is orange. Third row: salient colors in the image distract the regularized model from attending to the correct image regions. Fourth row: both the baseline and regularized models perform poorly on questions where the answer relates to a localized image region (i.e., inside a TV) as opposed to the global image. In these cases, the regularized model relies on generic visual features in the image in its inferences, while the baseline model relies on language priors.}
\label{fig:app:qualitative_failures}
\end{figure*}

\end{document}